# Latent Map Gaussian Processes for Mixed Variable Metamodeling

**Nicholas Oune**
Mechanical and Aerospace Engineering,
University of California, Irvine
Irvine, California, USA

**Ramin Bostanabad**[1]
Mechanical and Aerospace Engineering,
University of California, Irvine
Irvine, California, USA

**Abstract**

Gaussian processes (GPs) are ubiquitously used in sciences and engineering as metamodels. Standard GPs, however, can only handle numerical or quantitative variables. In this paper, we introduce latent map Gaussian processes (LMGPs) that inherit the attractive properties of GPs and are also applicable to mixed data which have both quantitative and qualitative inputs. The core idea behind LMGPs is to learn a continuous, low-dimensional latent space or manifold which encodes all qualitative inputs. To learn this manifold, we first assign a unique prior vector representation to each combination of qualitative inputs. We then use a low-rank linear map to project these priors on a manifold that characterizes the posterior representations. As the posteriors are quantitative, they can be directly used in any standard correlation function such as the Gaussian or Matern. Hence, the optimal map and the corresponding manifold, along with other hyperparameters of the correlation function, can be systematically learned via maximum likelihood estimation. Through a wide range of analytic and real-world examples, we demonstrate the advantages of LMGPs over state-of-the-art methods in terms of accuracy and versatility. In particular, we show that LMGPs can handle variable-length inputs, have an explainable neural network interpretation, and provide insights into how qualitative inputs affect the response or interact with each other. We also employ LMGPs in Bayesian optimization and illustrate that they can discover optimal compound compositions more efficiently than conventional methods that convert compositions to qualitative variables via manual featurization.

**Keywords:** Gaussian processes; emulation; metamodeling; mixed-variable optimization; computer experiments; and manifold learning.

---

[1] Corresponding author, raminb@uci.edu.

# 1 INTRODUCTION

Metamodeling (a.k.a. emulation, surrogate modeling, or supervised learning) of physical experiments or expensive simulations is critical for the development of research in many fields of science and engineering. As an example, consider the design of the airfoil shape for an aircraft wing. Many possible airfoil designs exist and testing each design, physically or via finite element (FE) simulations, could take minutes to possibly days. Metamodels, which are designed to mimic the input-output behavior in a system, make it possible to rapidly test airfoil designs due to their computationally inexpensive nature. Many metamodels have been developed over the past few decades. Some popular ones are based on Gaussian processes (GPs, aka Kriging) [1-11], neural networks (NNs) [12-17], and trees [18-20]. In this paper, we focus on GPs which are easy to train, quantify prediction uncertainty, and perform well with sparse datasets [2, 3, 21-27].

Standard methods for GP modeling assume all input variables are numerical because all covariance functions rely on a distance measure. However, in many real-world applications some features may be qualitative, or categorical, for which defining a distance measure is not straightforward. For example, consider the material design problem of identifying the optimal composition of the lacunar spinal family $XY_3^a Y_3^b Z_8$ with trivalent main group $X$, transition metal $Y$ and chalcogenide $Z$ ions [28]. The lacunar spinel family contains properties desirable for microelectronics, and the goal is to find the composition that maximizes phase stability and band gap tunability. In this design example, the inputs are all categorical and include elements for each site, e.g., either of $\{Al, Ga, In\}$ for the $X$ site. Since the differences such as $Al - Ga$ are not defined, GPs are not directly applicable to such problems. In this paper, we propose an efficient method to handle categorical variables via GPs.

The rest of the paper is organized as follows. Section 2 reviews standard methods for GP modeling. Section 3 summarizes existing techniques developed for handling qualitative inputs via GPs. Section 4 discusses our proposed strategy for training GPs on mixed datasets. Potential modifications to our approach are also discussed in this section. Section 5 reports the results from evaluating our method against state-of-the-art on a set of analytic and real-world problems. Section 6 concludes the paper with some final remarks.

## 2 GAUSSIAN PROCESS MODELING

In this section, we review how to fit GP models to a purely numerical training dataset whose inputs and outputs are denoted by $\boldsymbol{x} = [x_1, x_2, \ldots, x_{d_x}]^T$ and $y$, respectively. Assume the training data come from a realization of a Gaussian random process, $\eta(\boldsymbol{x})$, defined as the following:

$$\eta(\boldsymbol{x}) = \boldsymbol{f}(\boldsymbol{x})\boldsymbol{\beta} + \xi(\boldsymbol{x}),$$

where $\boldsymbol{f}(\boldsymbol{x}) = [f_1(\boldsymbol{x}), \ldots, f_h(\boldsymbol{x})]$ are a set of pre-determined parametric basis functions (e.g., $x_1^2 x_2$, $x_2^2 \sin(x_1)$, $\log(x_1 x_2)$, ...), $\boldsymbol{\beta} = [\beta_1, \ldots, \beta_h]^T$ are the unknown coefficients of the basis functions, and $\xi(\boldsymbol{x})$ is a zero-mean GP. Since $\xi(\boldsymbol{x})$ is zero-mean, it is completely characterized by its parameterized covariance function:

$$cov(\xi(\boldsymbol{x}), \xi(\boldsymbol{x}')) = c(\boldsymbol{x}, \boldsymbol{x}') = \sigma^2 r(\boldsymbol{x}, \boldsymbol{x}'), \tag{1}$$

where $\sigma^2$ is the process variance and $r(\cdot,\cdot)$ is a user-defined parametric correlation function. There are many types of correlation functions to choose from [1, 29-31], but the most common one is the Gaussian correlation function, which is defined as follows:

$$r(\boldsymbol{x}, \boldsymbol{x}') = exp\{-\sum_{i=1}^{d_x} 10^{\omega_i}(x_i - x_i')^2\} = exp\left((\boldsymbol{x} - \boldsymbol{x}')^T 10^{\boldsymbol{\Omega}}(\boldsymbol{x} - \boldsymbol{x}')\right), \tag{2}$$

where $\boldsymbol{\omega} = [\omega_1, \ldots, \omega_{d_x}]^T$, $-\infty < \omega_i < \infty$ are the roughness or scale parameters (in practice the ranges are limited to $-10 < \omega_i < 6$ to ensure numerical stability) and $\boldsymbol{\Omega} = diag(\boldsymbol{\omega})$. $\sigma^2$ and $\boldsymbol{\omega}$ are collectively referred to as the hyperparameters of the covariance function.

For GP emulation, point estimates of $\boldsymbol{\beta}, \boldsymbol{\omega}$, and $\sigma^2$ must be determined based on the data. These estimates can be found via either cross-validation (CV) or maximum likelihood estimation (MLE). Alternatively, Baye's rule can be applied to find posterior distributions of the hyperparameters if prior knowledge is available. In this paper, MLE is employed because it provides a high generalization power while minimizing the computational costs [1, 32]. MLE works by estimating $\boldsymbol{\beta}, \boldsymbol{\omega}$, and $\sigma^2$ such that they maximize the likelihood of the $n$ training data being generated by $\eta(\boldsymbol{x})$, that is:

$$[\widehat{\boldsymbol{\beta}}, \widehat{\sigma}, \widehat{\boldsymbol{\omega}}] = \underset{\boldsymbol{\beta},\sigma^2,\boldsymbol{\omega}}{\operatorname{argmax}} \; |2\pi\sigma^2 \boldsymbol{R}|^{-\frac{1}{2}} \times exp\left\{\frac{-1}{2}(\boldsymbol{y} - \boldsymbol{F}\boldsymbol{\beta})^T(\sigma^2 \boldsymbol{R})^{-1}(\boldsymbol{y} - \boldsymbol{F}\boldsymbol{\beta})\right\},$$

Or equivalently,

$$[\hat{\boldsymbol{\beta}}, \hat{\sigma}, \hat{\boldsymbol{\omega}}] = \underset{\boldsymbol{\beta}, \sigma^2, \boldsymbol{\omega}}{\text{argmin}} \ \frac{n}{2} log(\sigma^2) + \frac{1}{2} log(|\boldsymbol{R}|) + \frac{1}{2\sigma^2}(\boldsymbol{y} - \boldsymbol{F}\boldsymbol{\beta})^T \boldsymbol{R}^{-1}(\boldsymbol{y} - \boldsymbol{F}\boldsymbol{\beta}), \qquad (3)$$

where $log(\cdot)$ is the natural logarithm, $|\cdot|$ denotes the determinant operator, $\boldsymbol{y} = [y_{(1)}, \ldots, y_{(n)}]^T$ is an $n \times 1$ vector of outputs in the training data, $\boldsymbol{R}$ is the $n \times n$ correlation matrix with the $(i,j)^{th}$ element $\boldsymbol{R}_{ij} = r(\boldsymbol{x}_{(i)}, \boldsymbol{x}_{(j)})$ for $i, j = 1, \ldots, n$, and $\boldsymbol{F}$ is an $n \times h$ matrix with the $(k, l)^{th}$ element $\boldsymbol{F}_{kl} = f_l(\boldsymbol{x}_{(k)})$ for $k = 1, \ldots, n$ and $l = 1, \ldots, h$. By setting the partial derivatives with respect to $\boldsymbol{\beta}$ and $\sigma^2$ to zero, their estimates can be solved in terms of $\boldsymbol{\omega}$ as follows:

$$\hat{\boldsymbol{\beta}} = [\boldsymbol{F}^T \boldsymbol{R}^{-1} \boldsymbol{F}]^{-1} [\boldsymbol{F}^T \boldsymbol{R}^{-1} \boldsymbol{y}], \qquad (4)$$

$$\hat{\sigma}^2 = \frac{1}{n}(\boldsymbol{y} - \boldsymbol{F}\hat{\boldsymbol{\beta}})^T \boldsymbol{R}^{-1}(\boldsymbol{y} - \boldsymbol{F}\hat{\boldsymbol{\beta}}), \qquad (5)$$

Plugging these estimates into Eq. (3) and removing the constants yields:

$$\hat{\boldsymbol{\omega}} = \underset{\boldsymbol{\omega}}{\text{argmin}} \ nlog(\hat{\sigma}^2) + log(|\boldsymbol{R}|) = \underset{\boldsymbol{\omega}}{\text{argmin}} L. \qquad (6)$$

By minimizing $L$ (i.e., solving Eq. (6)), one can solve for $\hat{\boldsymbol{\omega}}$ and subsequently, obtain $\hat{\boldsymbol{\beta}}$ and $\hat{\sigma}^2$ using Eq. (4) and Eq. (5). While many heuristic global optimization methods exist such as genetic algorithms [33] and particle swarm optimization [34], gradient-based optimization techniques based on, e.g., the L-BFGS algorithm [35], are generally preferred due to their ease of implementation and superior computational efficiency [3, 29]. With gradient-based approaches, it is essential to start the optimization via numerous initial guesses to improve the chances of achieving global optimality.

After obtaining the hyperparameters via MLE, the following closed-form formula is used to predict the response at any $\boldsymbol{x}^*$:

$$\mathbb{E}[\boldsymbol{y}^*] = \boldsymbol{f}(\boldsymbol{x}^*)\hat{\boldsymbol{\beta}} + \boldsymbol{g}^T(\boldsymbol{x}^*) \boldsymbol{V}^{-1}(\boldsymbol{y} - \boldsymbol{F}\hat{\boldsymbol{\beta}}),$$

where $\mathbb{E}$ denotes expectation, $\boldsymbol{f}(\boldsymbol{x}^*) = [f_1(\boldsymbol{x}^*), \ldots, f_h(\boldsymbol{x}^*)]$, $\boldsymbol{g}(\boldsymbol{x}^*)$ is an $n \times 1$ vector with the $i^{th}$ element $c(\boldsymbol{x}_{(i)}, \boldsymbol{x}^*) = \hat{\sigma}^2 r(\boldsymbol{x}_{(i)}, \boldsymbol{x}^*)$, and $\boldsymbol{V}$ is the covariance matrix with the $(i,j)^{th}$ element $\hat{\sigma}^2 r(\boldsymbol{x}_{(i)}, \boldsymbol{x}_{(j)})$. The posterior covariance between the responses at the two inputs $\boldsymbol{x}^*$ and $\boldsymbol{x}'$ is:

$$cov(y^*, y') = c(\boldsymbol{x}^*, \boldsymbol{x}') - \boldsymbol{g}^T(\boldsymbol{x}^*) \boldsymbol{V}^{-1} \boldsymbol{g}(\boldsymbol{x}') + \boldsymbol{h}(\boldsymbol{x}^*)^T (\boldsymbol{F}^T \boldsymbol{V}^{-1} \boldsymbol{F})^{-1} \boldsymbol{h}(\boldsymbol{x}'),$$

where $\boldsymbol{h}(\boldsymbol{x}^*) = (\boldsymbol{f}(\boldsymbol{x}^*) - \boldsymbol{F}^T \boldsymbol{V}^{-1} \boldsymbol{g}(\boldsymbol{x}^*))$.

The above formulations can be easily extended to cases where the dataset is noisy. GPs can address noise and smoothen data by using a nugget or jitter parameter, $\delta$ [36]. As a result, $\boldsymbol{R}$ becomes $\boldsymbol{R}_\delta = \boldsymbol{R} + \delta \boldsymbol{I}_{n \times n}$ where $\boldsymbol{I}_{n \times n}$ is the identity matrix of size $n \times n$. If the nugget parameter is used, the estimated (stationary) noise variance in the data will be $\delta \hat{\sigma}^2$. GPs are also applicable to multi-response datasets by using, e.g., a separable covariance function [37-39] which replaces $\sigma^2$ with the matrix $\boldsymbol{\Sigma}$ whose off-diagonal elements represent the covariance between the corresponding responses at any fixed $\boldsymbol{x}$.

As the above formulations indicate, GP modeling relies on the correlation function, $r(\cdot,\cdot)$ in Eq. (2). $r(\cdot,\cdot)$ measures the correlation between the outputs at any two input locations as a function of the relative distance between those two inputs. Since the distance between categorical variables (such as gender, zip code, country, material coating type, etc.) cannot be directly defined, standard GP modeling techniques are not applicable to datasets that contain categorical variables. This issue is well established in the literature [40] and in the next section, we review the most common existing strategies that address it by reformulating the covariance function such that it can handle categorical variables.

## 3 EXISTING APPROACHES FOR HANDLING CATEGORICAL VARIABLES

Let us denote the categorical inputs by $\boldsymbol{t} = [t_1, \ldots, t_{d_t}]^T$ where the total number of distinct levels for categorical variable $t_i$ is $m_i$. For instance, $t_1 = \{92697, 92093\}$ and $t_2 = \{math, physics, chemistry\}$ are two categorical inputs that encode zip code ($m_1 = 2$ levels) and course subject ($m_2 = 3$ levels), respectively. Inputs for mixed (numerical and categorical) training data are collectively denoted by $\boldsymbol{w} = [\boldsymbol{x}; \boldsymbol{t}]$ which is a column vector of size $(d_x + d_t) \times 1$.

### 3.1 Unrestrictive Covariance (UC)

One popular strategy for GP modeling with categorical variables, introduced by Qian et al. [41], assumes a correlation function with the following form:

$$r(\boldsymbol{w}, \boldsymbol{w}') = \prod_{i=1}^{d_t} \tau_{l,l'}^i \times exp\{-(\boldsymbol{x} - \boldsymbol{x}')^T 10^{\boldsymbol{\Omega}} (\boldsymbol{x} - \boldsymbol{x}')\}, \qquad (7)$$

where $\tau_{l,l'}^i$ is a parameter that correlates levels $l$ and $l'$ of the variable $t_i$. That is, Eq. (7) assigns the correlation matrix $\boldsymbol{\tau}^i$ to the categorical variable $t_i$ where $\tau_{l,l'}^i$ serves as a distance metric between levels $l$ and $l'$ of $t_i$. Since $\boldsymbol{\tau}^i$ is a correlation matrix, it must be symmetric positive definite with unit diagonal elements. Hence, there are a total of $\sum_{i=1}^{d_t} m_i(m_i - 1)/2$ parameters that need to be estimated (in addition to $\boldsymbol{\Omega}$) in Eq. (7). Since the number of hyperparameters in the UC function increases quadratically, it is not applicable to problems where there are many levels or categorical variables. Even in simple problems, the constraints on $\boldsymbol{\tau}^i$ render the optimization of the log-likelihood function quite difficult. Additionally, Eq. (7) has limited generalization power. For example, as Deng et al. [42] point out, if $\tau_{l,l'}^i$ is estimated as 0 for any categorical variable, the entire correlation between two sample points, $r(\boldsymbol{w}, \boldsymbol{w}')$, reduces to 0.

## 3.2 Multiplicative Covariance (MC)

Multiplicative covariance function is a simplified version of the UC function [41] which assumes that for all $\boldsymbol{t} \neq \boldsymbol{t}'$:

$$\tau_{l,l'}^i = exp\{-(\theta_l^i + \theta_{l'}^i)\}, \qquad (8)$$

where $\theta_l^i > 0$ is a parameter associated with the $l^{th}$ level of categorical variable $t_i$. That is, Eq. (8) assigns a number to each level of each categorical variable and hence requires estimating a total of $\sum_{i=1}^{d_t} m_i$ parameters (in addition to $\boldsymbol{\Omega}$) during the MLE process. While the MC function is simpler to optimize than the UC function, it is quite inflexible [43]. To demonstrate this, consider a scenario where there is one categorical variable with four levels. Also, suppose that the response surfaces corresponding to $(i)$ levels 1 and 2 are highly correlated, $(ii)$ levels 3 and 4 are highly correlated, and $(iii)$ levels 1 and 2 are uncorrelated with those of levels 3 and 4. According to Eq. (8), we need to have $\theta_1 \approx \theta_2 \approx 0$ for the response surfaces for levels 1 and 2 to be highly correlated. With a similar reasoning, we have $\theta_3 \approx \theta_4 \approx 0$. However, for the response surfaces of levels 2 and 3 to be uncorrelated, $\theta_2 + \theta_3$ must be large, which cannot be true if both are close to zero.

## 3.3 Additive Gaussian Process (AGP)

The UC and MC strategies both assume a multiplicative covariance structure. Deng et al. [42] proposed a new additive covariance structure as follows:

$$cov(w(x,t), w'(x',t')) = \sum_{i=1}^{d_t} \sigma_i^2 \tau_{l,l'}^i r(x, x'|\boldsymbol{\omega}_i)$$

where $r(x, x'|\boldsymbol{\omega}_i)$ is the Gaussian correlation function as defined in Eq. (2) with correlation parameter vector $\boldsymbol{\omega}_i$ associated with qualitative factor $t_i$, $\sigma_i^2$ is the prior variance term for categorical variable $t_i$, and $\tau_{l,l'}^i$ has the same definition as in Eq. (8). According to Deng et al. [42], the AGP is more flexible than the UC function when there are multiple categorical variables. This is because the UC model assumes a fixed covariance structure over the numerical features, $x$, for all categorical variables while the additive structure does not. However, the AGP also has a few major limitations. For example, the optimization based on MLE involves estimating a total number of $(1 + d_x) \times d_t + \sum_{i=1}^{d_t} m_i(m_i - 1)/2$ parameters which rapidly increases as the dimensionality of the problem grows. Visualization of how the underling response surface changes within and across the categorical variables is also not straightforward with AGP.

### 3.4 Latent Variable Gaussian Process (LVGP)

Latent Variable Gaussian Process [44] is a recent work that handles categorical variables by learning a latent space of dimensionality $d_z$ for each categorical variable $t_i$ for $i = 1, ..., d_t$. In other words, the $m_i$ levels of $t_i$ are represented as $m_i$ points in the $i^{th}$ latent space. With this latent representation, the distance between any two points is defined. Hence, the latent points can be directly used in any standard correlation function such as the Gaussian:

$$r(w, w') = exp\left\{-\sum_{i=1}^{d_t} \|z^i(t_i) - z^i(t_i')\|_2^2 - (x - x')^T 10^{\Omega}(x - x')\right\}, \quad (9)$$

where $z^i(l) = [z_1^i(l), ..., z_{d_z}^i(l)]^T$ is the latent space point corresponding to level $l$ (for $l = 1, ..., m_i$) of the qualitative factor $t_i$ and $\|\cdot\|_2$ denotes the Euclidean 2-norm. With this formulation, all the latent points (along with $\boldsymbol{\omega}$) can be found via MLE as described in Section 2 where Eq. (2) must be replaced via Eq. (9). Zhang et. al. [44] recommend using a $2D$ latent space for each categorical variable where three constraints are imposed to ensure translation and rotation invariances in each $2D$ space. Thus, fitting an LVGP model involves estimating $d_x + \sum_{i=1}^{d_t}(2m_i - 3)$ parameters.

Zhang et al. [44] show that LVGP consistently outperforms previously mentioned strategies in a wide range of problems. This superior performance is primarily because $(i)$ in many real-world scenarios, categorical variables represent underlying numerical features whose collective effects can be captured in the learned $2D$ latent space, and $(ii)$ the correlation function in Eq. (9) provides a much more versatile reformulation and does not impose any a priori relation between the categorical variables.

## 4 LATENT MAP GAUSSIAN PROCESS (LMGP)

Our proposed approach, similar to LVGP, involves mapping categorical variables to some latent points. However, there are two key differences between LMGP and LVGP. Firstly, instead of directly estimating the latent positions, LMGP learns a linear transformation that maps a prior representation of the categorical variables to the latent space. Secondly, LMGP uses a single latent space while LVGP uses a unique latent space for each categorical variable. As argued below and shown in Section 5, these differences make LMGP a more versatile and accurate metamodel than LVGP.

In Section 4.1 we provide the motivation for LMGP and the technical details. In Section 4.2, we draw some connections between LMGP and some other concepts (NNs, sufficient dimension reduction, and active subspaces) and also introduce an extension that enables LMGP to handle variable-length (or conditional) categorical inputs.

### 4.1 Latent Space Representation

Mapping categorical variables onto a latent space has a strong justification because in all physical systems with such inputs, there exist some underlying numerical features that characterize the levels of each categorical variable. For example, consider the lacunar spinal family design problem discussed in the introduction. The differences between the elements in the periodic table can be captured through some numerical features such as atomic number, atomic mass, number of valence electrons, and more. While there may be many numerical features that characterize the level-wise differences within and across the categorical variables, these features are generally highly correlated or may have little effect on the response of interest [12, 45]. Hence, a low dimensional latent space is generally sufficient to capture the effect of the underlying numerical features. Using latent spaces is also advantageous in cases where the underlying numerical features are not (completely) known or cannot be directly observed or measured. These and similar arguments underlie the widespread use of latent representations in deep NNs which, unlike LMGP, are generally constructed using big data.

LMGP begins with an initial latent space representation of the categorical inputs. This prior representation is then projected to a lower dimensional space via a linear map which is learned via MLE, see **Figure 1**. In particular, we first assign a unique vector (i.e., a prior representation) to each combination of the categorical variables. Then, we use matrix multiplication to map each of these unique vectors to a point in a latent space of dimensionality $d_z$:

$$z(t) = \zeta(t)A,$$

where $t$ is a particular combination of the categorical variables, $z(t)$ is the $1 \times d_z$ posterior latent representation of $t$, $\zeta(t)$ is the $1 \times \sum_{i=1}^{d_t} m_i$ unique prior vector representation of $t$, and $A$ is a $\sum_{i=1}^{d_t} m_i \times d_z$ mapping matrix that maps $\zeta(t)$ to $z(t)$. These points can now be directly inserted into any standard correlation function such as the Gaussian:

$$r(w, w') = exp\{-\|z(t) - z(t')\|_2^2 - (x - x')^T 10^{\Omega}(x - x')\}. \quad (10)$$

Finally, we optimize $A$ simultaneously with $\Omega = diag(\omega)$ via MLE:

$$[\widehat{\omega}, \widehat{A}] = \underset{\omega, A}{\operatorname{argmin}} \ nlog(\hat{\sigma}^2) + log(|R|),$$

where $R$ and $\hat{\sigma}^2$ are now functions of both $\omega$ and $A$. When a $2D$ latent space is used ($d_z = 2$), which we do in this paper, three constraints can be applied to the posterior latent positions to ensure rotation and translation invariance of the learned representation. Denoting the horizontal and vertical axes of this posterior space by $z_1$ and $z_2$, respectively, these constraints are: $(i)$ The first latent position is located at the origin ($z_1 = z_2 = 0$), $(ii)$ the second latent position has $z_1 \geq 0$ and $z_2 = 0$, and $(iii)$ the third latent position has $z_2 \geq 0$.

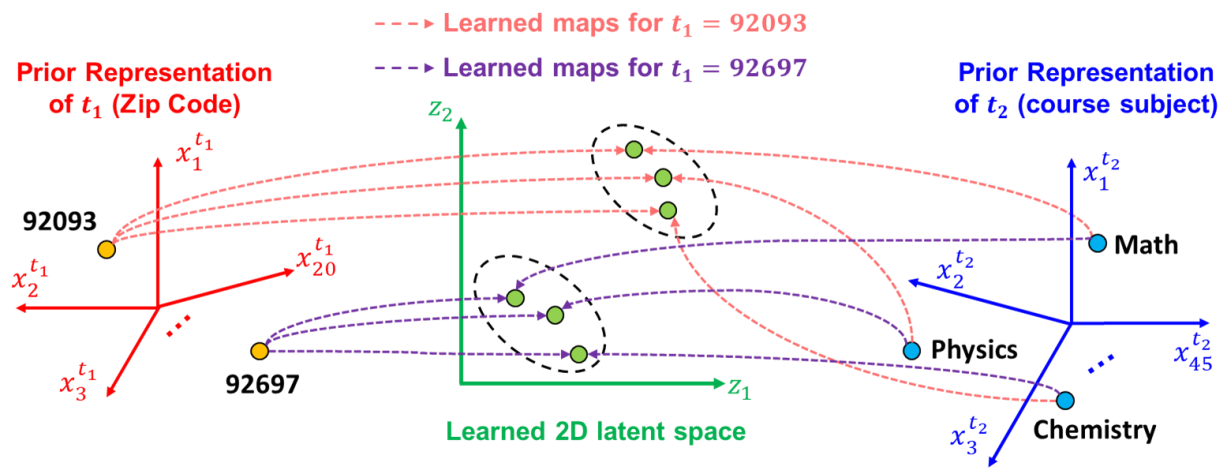

**Figure 1 Learning latent space via LMGP:** The high-dimensional prior representations of categorical variables are mapped into a $2D$ latent space where the mapping is learnt via MLE. The mapped are colored based on the levels of $t_1$. In this illustrative example, changing the level of $t_1$ affects the latent positions more and hence the response is more sensitive to $t_1$ than $t_2$.

While $\boldsymbol{A}$ is learned via MLE based on some training data, the prior representations, $\boldsymbol{\zeta}(\boldsymbol{t})$, are user defined and can affect the performance of LMGP. We propose two strategies for defining $\boldsymbol{\zeta}(\boldsymbol{t})$. One method, which we call the random initialization, is to define $\boldsymbol{\zeta}(\boldsymbol{t})$ as a $1 \times \sum_{i=1}^{dt} m_i$ vector of random values ranging from, e.g., 0 to 1 (other ranges can be used which will result in larger/smaller estimates for the elements of $\boldsymbol{A}$). For instance, consider the example in Section 3 where $t_1 = \{92697, 92093\}$ and $t_2 = \{math,\ physics, chemistry\}$. With random initialization, $\boldsymbol{\zeta}(\boldsymbol{t})$ for each combination of levels of the categorical variables is defined as follows:

$$\begin{bmatrix} \{92697, math\} \\ \{92697, physics\} \\ \{92697, chemistry\} \\ \{92093, math\} \\ \{92093, physics\} \\ \{92093, chemistry\} \end{bmatrix} \rightarrow \begin{bmatrix} \boldsymbol{\zeta}(1,1) \\ \boldsymbol{\zeta}(1,2) \\ \boldsymbol{\zeta}(1,3) \\ \boldsymbol{\zeta}(2,1) \\ \boldsymbol{\zeta}(2,2) \\ \boldsymbol{\zeta}(2,3) \end{bmatrix} = \boldsymbol{\chi}_{6\times 5}, \qquad \chi_{ij} \sim Uni(0,1)$$

where $\boldsymbol{\zeta}(a, b)$ is the unique vector representation when the first and second categorical variables are at levels $a$ and $b$, respectively, and $\boldsymbol{\chi}$ is a matrix whose elements are independent and identically distributed (IID) random numbers that follow a standard uniform distribution. While $\boldsymbol{\chi}$ is random and completely changes[2] each time we fit an LMGP to a particular dataset, our studies indicate that the resulting latent positions (and hence the accuracy of LMGP) are not affected. This consistency in posterior representation is provided by $\boldsymbol{A}$ whose elements are estimated via MLE. Furthermore, to reduce computational costs (esp. in very high dimensional problems), one can reduce the number of columns of $\boldsymbol{\chi}$ which will reduce the number of rows of $\boldsymbol{A}$. However, we found through testing that this would be at the expense of potential reduction in prediction performance.

The second initialization strategy is to use a grouped one-hot encoded vector for $\boldsymbol{\zeta}(\boldsymbol{t})$ that consists of 1s and 0s. In a $1 - 0$ vector representation, the 1s correspond to the levels used for each categorical variable while the 0s correspond to the rest of the levels. By applying this approach to the zip code-course subject example, $\zeta(t)$ is obtained as follows:

$$\begin{bmatrix} \{92697, math\} \\ \{92697, physics\} \\ \{92697, chemistry\} \\ \{92093, math\} \\ \{92093, physics\} \\ \{92093, chemistry\} \end{bmatrix} \rightarrow \begin{bmatrix} \zeta(1,1) \\ \zeta(1,2) \\ \zeta(1,3) \\ \zeta(2,1) \\ \zeta(2,2) \\ \zeta(2,3) \end{bmatrix} = \begin{bmatrix} 1 & 0 & 1 & 0 & 0 \\ 1 & 0 & 0 & 1 & 0 \\ 1 & 0 & 0 & 0 & 1 \\ 0 & 1 & 1 & 0 & 0 \\ 0 & 1 & 0 & 1 & 0 \\ 0 & 1 & 0 & 0 & 1 \end{bmatrix}_{6\times 5}$$

---

[2] Assuming the random number generator seed is not fixed

In our studies, the $1-0$ representation consistently outperformed the random representation based on the performance of LMGP on test data. It also resulted in better structured latent positions that more clearly demonstrate the relations between the categorical variables and their relative effect on the response. These favorable properties are because the $1-0$ representation acts as an informative prior that helps LMGP in distinguishing the interactions between categorical variables and their levels. However, the random vector representation provides an uninformative prior where $\zeta(t)$ is not generated based on any knowledge of the categorical variable levels.

To verify that LMGP is actually utilizing knowledge of the levels used for each categorical variable, we compared its prediction performance in two scenarios: standard LMGP (as described above with $1-0$ representation) and LMGP that lumps all categorical variables into a single new one where each level corresponds to a set of levels for the original categorical variable. Consider the zip code-course subject example again. By combining the two categorical variables into a single new categorical variable, $\zeta(t)$ becomes a diagonal matrix as shown below:

$$\begin{bmatrix} \{92697, math\} \\ \{92697, physics\} \\ \{92697, chemistry\} \\ \{92093, math\} \\ \{92093, physics\} \\ \{92093, chemistry\} \end{bmatrix} \rightarrow \begin{bmatrix} \zeta(1) \\ \zeta(2) \\ \zeta(3) \\ \zeta(4) \\ \zeta(5) \\ \zeta(6) \end{bmatrix} = \begin{bmatrix} 1 & 0 & 0 & 0 & 0 & 0 \\ 0 & 1 & 0 & 0 & 0 & 0 \\ 0 & 0 & 1 & 0 & 0 & 0 \\ 0 & 0 & 0 & 1 & 0 & 0 \\ 0 & 0 & 0 & 0 & 1 & 0 \\ 0 & 0 & 0 & 0 & 0 & 1 \end{bmatrix}_{6\times 6}$$

Because the categorical variables are combined into a single new categorical variable, the levels used for each categorical variable is unknown to LMGP. Thus, poorer prediction performance is expected. Through testing, we found that standard LMGP consistently outperformed LMGP with the categorical variables combined. This implies that the prior knowledge of the levels for each categorical variable is assisting LMGP with discovering a more representative latent position structure.

We argue that LMGP is a more suitable approach to latent space learning than LVGP because of the following four key reasons. Firstly, LMGP provides a systematic mechanism to embed prior knowledge (the $1-0$ vector representation in our case) into the training process while LVGP directly estimates the latent positions. Our mechanism greatly improves the optimization process which, in turn, results in models with higher predictive power. Secondly, mapping all possible $\boldsymbol{t}$ vectors to a single latent space (as opposed to having a latent space for each categorical variable) allows the user to analyze and visualize the interactions across the categorical variables. Thirdly, while LMGP requires estimating more hyperparameters than LVGP, it achieves a more aggressive dimensionality reduction (the total number of hyperparameters in LMGP and LVGP are $d_x + d_z \times \sum_{i=1}^{q} m_i$ and $d_x + \sum_{i=1}^{d_t}(2m_i - 3)$, respectively). This is because all the latent positions in LMGP are enforced to lie on a single latent space (aka manifold [46]) while LVGP uses a manifold for each categorical variable. Lastly, LMGP avoids non-identifiability issues that LVGP encounters: As we show in Section 5.1, when LVGP is trained on noisy data where one or more of the categorical variables have negligible effect on the response, the latent positions cannot be robustly estimated because their effect on the correlation function is akin to that of the nugget parameter. We demonstrate some of these remarks below and the rest in Section 5.

We now use the borehole function [47] to demonstrate some of the nice properties of LMGP. The borehole function has eight inputs and is ubiquitously used to assess the performance of surrogates. It is defined as:

$$y = \frac{2\pi T_u(H_u - H_l)}{\ln\left(\frac{r}{r_\omega}\right)\left(1 + \frac{2LT_u}{ln\left(\frac{r}{r_\omega}\right)r_\omega{}^2 K_\omega} + \frac{T_u}{T_l}\right)}. \tag{11}$$

The inputs of the borehole function are all quantitative. To have mixed variables, we convert $T_l$, $L$, and $K_w$ into categorical variables with $5, 3,$ and $3$ levels, respectively. Each level corresponds to a distinct numerical value *unknown* to LMGP (see **Table 11** in the Appendix for details). The latent space positions estimated via LMGP are demonstrated in **Figure 2** where the legend shows the combination of levels (the triplets belong to $T_l$, $L$, and $K_w$, respectively) that corresponds to a point in the $2D$ latent space. Notice that the range of the axes is quite different and that the estimated latent positions are structured on a grid with eight vertical and five horizontal lines. On the vertical lines only the level of the first categorical variable changes (to see this, locate the markers with the same shape in the legend) while on the horizontal lines either the level of $L$ or $K_w$ changes. This figure suggests that the underlying function, while having 3 categorical variables, only depends on two hidden features. Furthermore, the range of the axes indicates that one of these hidden features affects the response, i.e., $y$ in Eq. (11), more than the other. This relative importance of the two features is deduced from the term $-\|\mathbf{z}(t) - \mathbf{z}(t')\|_2^2$ in Eq. (10): in **Figure 2**, the hidden feature that is encoded by the horizontal axis has more variations and thus contributes more to this term. The higher contribution indicates that this feature affects the response more than the feature encoded by the vertical axis.

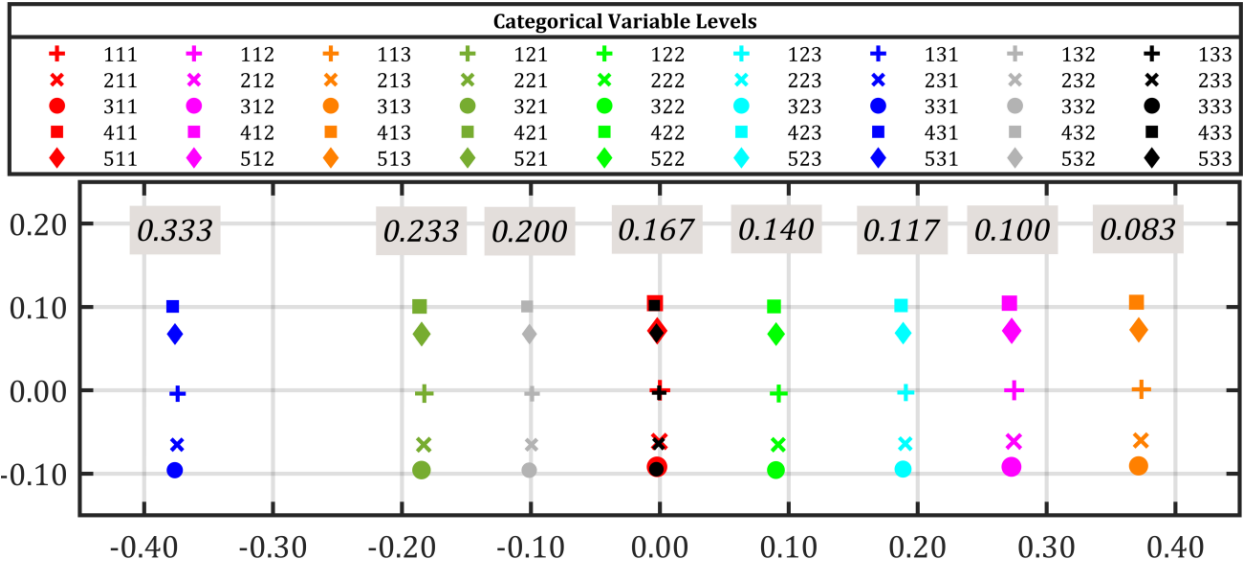

**Figure 2 Latent positions learned via LMGP for the borehole function:** Each latent position corresponds to a set of levels for each categorical variable in the borehole function. The first, second, and third categorical variables correspond to $T_l$, $L$, and $K_w$, respectively. The points with the same level of $T_l$ are structured approximately on a horizontal line. The points with the same levels of $L$ and $K_w$ are structured approximately on a vertical line. The underlying numerical value of $L/K_w$ is indicated in the grey box on top of each vertical line.

To relate the above insights with the underlying function, we rewrite the borehole function as:

$$y = \frac{C_1}{C_2\left(1 + C_3\frac{L}{K_w} + \frac{C_4}{T_l}\right)} = f\left(\frac{L}{K_w}, T_l, C_1, C_2, C_3, C_4\right) \tag{12}$$

where the numerical variables are all lumped to $C_1, C_2, C_3$ and $C_4$. Eq. (12) clearly shows that the three original categorical variables can be compressed to two variables, namely, $L/K_w$ and $T_l$. In fact, these two variables are the hidden features that LMGP learns purely based on the data. That is, in **Figure 2**, one axis encodes $T_l$ while the other axis encodes $L/K_w$ (note that the latter is a nonlinear function of the original variables $L$ and $K_w$). Next, we find the correspondence between the axes of the latent space with $L/K_w$ and $T_l$.

**Table 1** lists the underlying numerical value of $L/K_w$ for each combination of levels for $L$ and $K_w$. By matching these numbers with the latent points in **Figure 2** it can be seen that each vertical line is associated with a unique number and that these numbers monotonically decrease from left to right. That is, the left- and right-most latent positions have, respectively, the largest ($L/K_w = 0.333$) and the smallest ($L/K_w = 0.083$) values. Notice that the ratio $L/K_w$ is 0.167 when $L$ and $K_w$ both have a level of either 1 or 3 (see the first and last columns of Table 1). This situation is also accurately reflected in the latent space where the corresponding latent positions overlap (see the black and red markers in **Figure 2**). The preceding discussions highlight that LMGP accurately discovers the latent feature that captures the collective effects of both $L$ and $K_w$. This latent feature is encoded by the horizontal axis in **Figure 2** and thus the vertical axis encodes $T_l$.

**Table 1 Underlying numerical values of $L/K_w$:** Values are reported for each combination of levels for $L$ and $K_w$.

| Level of $L$ | 1 | 1 | 1 | 2 | 2 | 2 | 3 | 3 | 3 |
|---|---|---|---|---|---|---|---|---|---|
| Level of $K_w$ | 1 | 2 | 3 | 1 | 2 | 3 | 1 | 2 | 3 |
| $L/K_w$ | 0.167 | 0.100 | 0.083 | 0.233 | 0.140 | 0.117 | 0.333 | 0.200 | 0.167 |

While the latent representation along the horizontal axis in **Figure 2** is consistent with the ratio $L/K_w$, the same is not true for $T_l$. The underlying numerical values of $T_l$ are organized in ascending order (they are $[10, 30, 100, 200, 500]$, see **Table 11**). So, a monotonically ascending/descending order is expected for the levels of $T_l$ in the latent space (The latent axis can have either an opposite or similar ordering as the underlying numerical values. In case of the horizontal axis, the positive direction is aligned with a reduction in $L/K_w$). That is, the levels of $T_l$ on the horizontal lines from top to bottom in **Figure 2** should be either $[1, 2, 3, 4, 5]$ or $[5, 4, 3, 2, 1]$. Instead, we see that the corresponding levels of $T_l$ are ordered as 3, 2, 1, 5, and 4 (from bottom to top on each horizontal line). To better understand why LMGP seems to discover a sub-optimal latent representation for $T_l$, we employ Sobol sensitivity analysis [48].

Sobol sensitivity analysis is a method used for analyzing each input's total contribution to the output variance given the range of the inputs. The input's total contribution to the output variance can be decomposed into two parts: variance from each individual input and variance from interactions among inputs. **Table 2** lists each input's total contribution to the output variance, referred to as the "total-effect index", for the borehole function in Eq. (11). The ranges used for the numerical features are described in **Table 11**, and the ranges used for the categorical variables are the maximum and minimum underlying numerical values which are also listed in **Table 11**.

**Table 2** quantitatively indicates that $T_l$ has a much smaller total-effect index than $L$ or $K_w$. In fact, the extremely small sensitivity index of $T_l$ signals that it almost has no effect on the response. This behavior is the reason that LMGP cannot find the correct latent representation for the varying levels of $T_l$. The sensitivity indices in **Table 2** also explain why the range of the axes in **Figure 2** are quite different: since $T_l$ negligibly affects the response, its contribution to the correlation function (through the term $-\|\boldsymbol{z}(\boldsymbol{t}) - \boldsymbol{z}(\boldsymbol{t'})\|_2^2$ in Eq. (10)) should be small which, in turn, requires small vertical distance between the latent points in **Figure 2**.

**Table 2 Total-effect index:** The total-effect index is a metric that defines each input's total (individually and through interaction) contribution to the output variance. Unlike $L$ and $K_w$, $T_l$ almost has no effect on the variability of the response which makes it difficult to encode it in the latent space.

|  | $T_u$ | $H_u$ | $H_l$ | $r$ | $r_w$ | $T_l$ | $L$ | $K_w$ |
|---|---|---|---|---|---|---|---|---|
| Total-effect Index | 0.0000 | 0.0463 | 0.0465 | 0.0000 | 0.7445 | 0.0001 | 0.1290 | 0.1177 |

## 4.2 Discussions and Extensions

In this section, we first elaborate on the connections between NNs and LMGPs which can lead to further developments of LMGPs. Then, we compare LMGPs with sufficient dimension reduction and active subspaces. Finally, we discuss how to handle variable-length (or conditional) inputs with LMGPs.

### 4.2.1 Neural Network Interpretation of LMGP

LMGP can be perceived as an NN that encodes the categorical variables to a latent space as shown in **Figure 3**. In this particular NN architecture, the latent space mapping in LMGP corresponds to a single hidden layer with linear activation functions and no bias terms (due to MLE's translation invariance). For this hidden layer, $\zeta(t)$ is the input, $z(t)$ is the output, and $A$ represents the neural network weights.

With this interpretation, we can extend LMGP in a few ways. First, we can increase the number of hidden layers and their neurons to improve the learning capacity at the expense of increasing the number of hyperparameters that must be estimated. Second, we can use a nonlinear activation function (e.g., sigmoid, swish, or tangent hyperbolic) instead of a linear one. In our studies, we have observed marginal changes when testing the second idea but the integration of the two ideas may be more effective.

### 4.2.2 Active Subspaces and Sufficient Dimension Reduction

Active subspace methods use the gradient information for dimensionality reduction [49]. Assuming the gradient of the function is available at training points, one starts by rotating the input space (i.e., a linear transformation via singular value decomposition, SVD) to separate the directions based on their variability. The input space is then projected to the directions where the most variability is observed. Finally, surrogate modeling (with GPs or any other method) is done at this lower dimensional space which is called the active subspace of the underlying function.

LMGP is similar to active subspace methods in that they both rely on linear transformations. However, there are some fundamental differences between the two. First, unlike LMGP, active subspaces are not applicable to categorical data as the derivates are not defined. Second, while both methods reduce dimensions through some linear transformations, the underlying mechanism behind LMGP is different because it is supervised, relies on MLE (rather than SVD), and does not require gradient information (it is noted that some active subspace methods also do not require gradients, for example if a GP is used for metamodeling [50]).

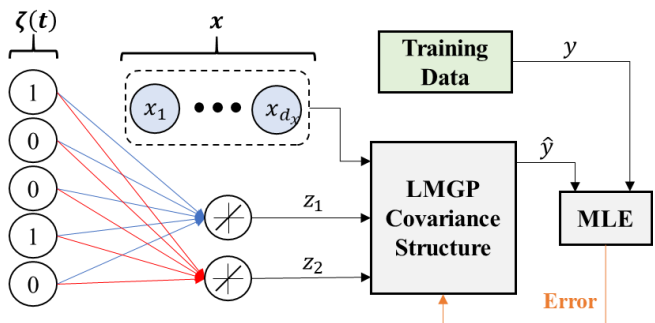

**Figure 3 Neural network interpretation of LMGP with a 2$D$ latent space:** Categorical data, $t$, are converted to prior vector representations, $\zeta(t)$, and fed into the network's hidden layer that maps $\zeta(t)$ to $z$ using linear activation functions and no bias. The LMGP covariance structure then uses $x$ and $z$ as inputs to approximate $y$.

Similar to active subspace methods, the core idea behind sufficient dimension reduction is to build a surrogate using a lower dimensional input space that is constructed via a linear transformation (SVD) of the original input space [51-53]. Sufficient dimension reduction does unsupervised dimension reduction (the responses may be used to slice the covariance matrices though) and cannot handle categorical inputs.

### 4.2.3 Variable-Length Categorical Inputs

Both standard LMGP and LVGP cannot accept variable-length inputs. However, LMGP can be easily modified to handle variable-length inputs. We demonstrate this using the zip code-course subject example where $t_1 = \{92697, 92093\}$ and $t_2 = \{math,\ physics, chemistry\}$. Assume that when $t_1 = 92093$, $t_2$ is no longer an input, i.e., the system's response is independent of $t_2$ if $t_1 = 92093$. This conditional situation is illustrated in **Table 3** where $NaN$ indicates that the categorical variable is not an input (or if it is an input, it does not affect the system's response).

**Table 3 Combinations of levels for the variable-length example:** All combinations of levels for the first and second categorical variable are listed. When the level is $NaN$, that categorical variable is not an input given the level of the other categorical variable.

| First Categorical Variable | Second Categorical Variable |
|---|---|
| 1 | 1 |
| 1 | 2 |
| 1 | 3 |
| 2 | $NaN$ |

To make LMGP compatible with variable-length categorical inputs, only the prior vector representations need to be adjusted. Applying the $1 - 0$ representation described in Section 4.1 to the current example results in:

$$\begin{bmatrix} \zeta(a=1,b=1) \\ \zeta(a=1,b=2) \\ \zeta(a=1,b=3) \\ \zeta(a=2,NaN) \end{bmatrix} = \begin{bmatrix} 1 & 0 & 1 & 0 & 0 \\ 1 & 0 & 0 & 1 & 0 \\ 1 & 0 & 0 & 0 & 1 \\ 0 & 1 & NaN & NaN & NaN \end{bmatrix}$$

where $\boldsymbol{\zeta}(a,b)$ is the unique vector representation when the first and second categorical variables are at levels $a$ and $b$, respectively Because $NaN$ is not a valid value it must be replaced with a number. For this, we propose two potential approaches. One strategy is to replace $NaN$ values with IID random numbers so $\boldsymbol{\zeta}(t)$ for each combination of levels of the categorical variables becomes:

$$\begin{bmatrix} \zeta(a=1,b=1) \\ \zeta(a=1,b=2) \\ \zeta(a=1,b=3) \\ \zeta(a=2,NaN) \end{bmatrix} = \begin{bmatrix} 1 & 0 & 1 & 0 & 0 \\ 1 & 0 & 0 & 1 & 0 \\ 1 & 0 & 0 & 0 & 1 \\ 0 & 1 & \chi & \chi & \chi \end{bmatrix}, \quad \chi \sim Uni(0,1)$$

Another strategy is to simply replace all $NaN$ values with 0. In Section 5.3, we see through testing that both strategies yield very similar performance.

## 5  RESULTS

In this section, we compare the performance of LMGP against LVGP and also apply LMGP to two variable-length problems. We do not compare LMGP with the other methods reviewed in Section 3 as LVGP is shown to consistently outperform them [44]. Both algorithms are coded in Matlab and leverage continuation [29] to estimate the optimum nugget variance. More algorithmic details on both LMGP and LVGP are provided in the Appendix, see Section 8.1.

In Section 5.1, we compare LMGP to LVGP using six analytical functions with various sample sizes, noise levels, and number of categorical variables. In Section 5.2, we apply both methods to two real-world datasets and in Section 5.3 we analyze the performance of LMGP on handling variable-length categorical inputs. Finally, in Section 5.4 we use LMGP in Bayesian optimization for identifying the compound composition that maximizes the bulk modulus.

### 5.1  Analytical Functions

**Table 4** summarizes the analytical functions used for comparing LMGP to LVGP. Since these functions only have numerical variables, we modify them by converting a few numerical features into categorical features. That is, each level of the categorical variables corresponds to an underlying numerical value unknown to LMGP and LVGP. The underlying numerical values for the categorical variable levels for each analytical function are listed in the Appendix, see Section 8.2. These analytical functions are chosen as they have a wide range of dimensionality and degree of nonlinearity. Additionally, the conversion of numerical variables to categorical ones, allows to have multiple categorical variables with many levels, see **Table 5**.

**Table 4 List of analytical functions:** The functions possess a wide range of dimensionality and complexity. When emulating each function, we use training datasets of sizes 100, 200, 300, and 400. We also add IID normal noise to both training and test data. Three noise variances are considered for each function. The smallest variance is 0 in all cases while the other two depend on the range of the function.

| ID-Name [Ref] | Function |
|---|---|
| 1 − OLT Circuit [54] | $$y = \frac{(V_{b1} + 0.74)\beta(R_{c2} + 9) + 11.35 R_f}{\beta(R_{c2} + 9) + R_f} + \frac{0.74 R_f \beta(R_{c2} + 9)}{\left(\beta(R_{c2} + 9) + R_f\right) R_{c1}}$$ $$V_{b1} = \frac{12 R_{b2}}{R_{b1} + R_{b2}}$$ |
| 2 − Piston Simulator [54] | $$y = 2\pi \sqrt{\frac{M}{k + S^2 \frac{P_0 V_0 T}{T_0 V^2}}}$$ $$V = \frac{S}{2k}\sqrt{A^2 + 4k\frac{P_0}{T_0}T}, \quad A = P_0 S + 19.62 M - \frac{k V_0}{S}$$ |
| 3 − Borehole [47] | $$y = \frac{2\pi T_u (H_u - H_l)}{\ln\left(\frac{r}{r_\omega}\right)\left(1 + \frac{2 L T_u}{\ln\left(\frac{r}{r_\omega}\right) r_\omega^2 K_\omega} + \frac{T_u}{T_l}\right)}$$ |
| 4 − Effective Potential [55] | $$y = 100 * \frac{9}{2} x_9 \varepsilon_m^2 + \frac{x_8 x_{10}}{1 + x_7}\left[\frac{\varepsilon_{eq}}{x_{10}}\right]^{1+x_7}$$ $$\boldsymbol{\varepsilon} = \begin{pmatrix} x_1 & x_6 & x_5 \\ x_6 & x_2 & x_4 \\ x_5 & x_4 & x_3 \end{pmatrix}, \quad \varepsilon_m = \frac{1}{3} Tr(\boldsymbol{\varepsilon}), \varepsilon_d = \boldsymbol{\varepsilon} - \varepsilon_m \mathbf{1}, \varepsilon_{eq} = \sqrt{\frac{2}{3}(\varepsilon_d : \varepsilon_d)}$$ |
| 5 − Wing Weight [56] | $$y = 0.036 S_\omega^{0.758} W_{f\omega}^{0.0035} \left(\frac{A}{\cos^2(\Lambda)}\right)^{0.6} q^{0.006} \lambda^{0.04} \left(\frac{100 t_c}{\cos(\Lambda)}\right)^{-0.3} (N_z W_{dg})^{0.49} + S_\omega W_p$$ |
| 6 − Custom Function [57] | $$y = 4(x_1 - 2 + 8 x_2 - 8 x_2^2)^2 + (3 - 4 x_2)^2 + 16\sqrt{x_3 + 1}(2 x_3 - 1)^2 + \sum_{i=4}^{8} i \ln\left(1 + \sum_{j=3}^{i} x_j\right)$$ |

**Table 5 Input descriptions:** The numerical (in red) and categorical (in blue) inputs, their ranges, and the number of level combinations, $b_t$, are listed for each function.

| ID | Variables (Numerical, Categorical) | Min, Max | $b_t$ |
|---|---|---|---|
| 1 | $R_{b1}, R_{b2}, R_f, R_{c1}, R_{c2}, \beta$ | $[1, 50, 1, 1.2, 0.01, 1]$, $[3, 70, 3, 2.5, 5, 3]$ | 27 |
| 2 | $M, S, V_0, k, P_0, T, T_0$ | $[1, 1, 1, 2000, 2E5, 10, 10]$, $[3, 3, 3, 3000, 1.5E6, 500, 760]$ | 27 |
| 3 | $T_u, H_u, H_l, r, r_w, T_l, L, K_w$ | $[100, 990, 700, 100, 0.05, 1, 1, 1]$, $[1000, 1110, 820, 1E4, 0.15, 5, 3, 3]$ | 45 |
| 4 | $x_1, x_2, x_3, x_4, x_5, x_6, x_7, x_8, x_9, x_{10}$ | $[0, 0, 0, 0, 0, 0, 1, 1, 1, 1]$, $[1, 1, 1, 1, 1, 1, 5, 5, 5, 5]$ | 625 |
| 5 | $S_w, W_{fw}, A, \Lambda, q, \lambda, t_c, N_z, W_{dg}, W_p$ | $[1, 1, 6, -10, 16, 0.5, 1, 2.5, 1, 0.025]$, $[3, 3, 10, 10, 45, 1, 3, 6, 3, 0.08]$ | 81 |
| 6 | $x_1, x_2, x_3, x_4, x_5, x_6, x_7, x_8$ | $[0, 0, 1, 1, 0, 0, 0, 1]$, $[1, 1, 6, 4, 1, 1, 1, 3]$ | 72 |

For testing, we compare the performance across various training dataset sizes (ranging between 100 to 400 samples) and noise levels, which varies based on the range of the analytical function. After fitting LMGP and LVGP, we then evaluate the mean square error (MSE) across 10,000 test samples. Training and validation input samples are (for both quantitative and categorical variables) generated via Sobol sequence [58-60]. To account for randomness and measure consistency, the training and validation process is repeated 10 times with a new dataset each time.

**Figure 4** summarizes the results and indicates that LMGP consistently outperforms LVGP across all noise levels for large training datasets. In particular, LMGP achieves a test MSE on noisy data that is very close to the noise variance which indicates that it is able to extract as much information from the data as possible. With small datasets, LVGP sometimes outperforms LMGP and is more robust. We believe this is due to the fact that LMGP has more hyperparameters than LVGP and hence needs more data. Additionally, we note that in two cases (borehole and wing weight functions), increasing the training dataset size decreases LVGP's performance which is unintuitive because the performance is generally expected to improve once more data are used in training. This unintuitive behavior of LVGP is due to its failure in distinguishing noise from variables that negligibly affect the response. For instance, in the borehole function, one of the categorical variables, $T_l$, insignificantly affects the response, as evidenced by its Sobol index in **Table 2**. In this case, LVGP is mistakenly interpreting the variations in $y$ that are rooted in $T_l$ to be due to noise. This non-identifiability issue is not seen in LMGP as it is able to distinguish noise from categorical variables that marginally affect the response.

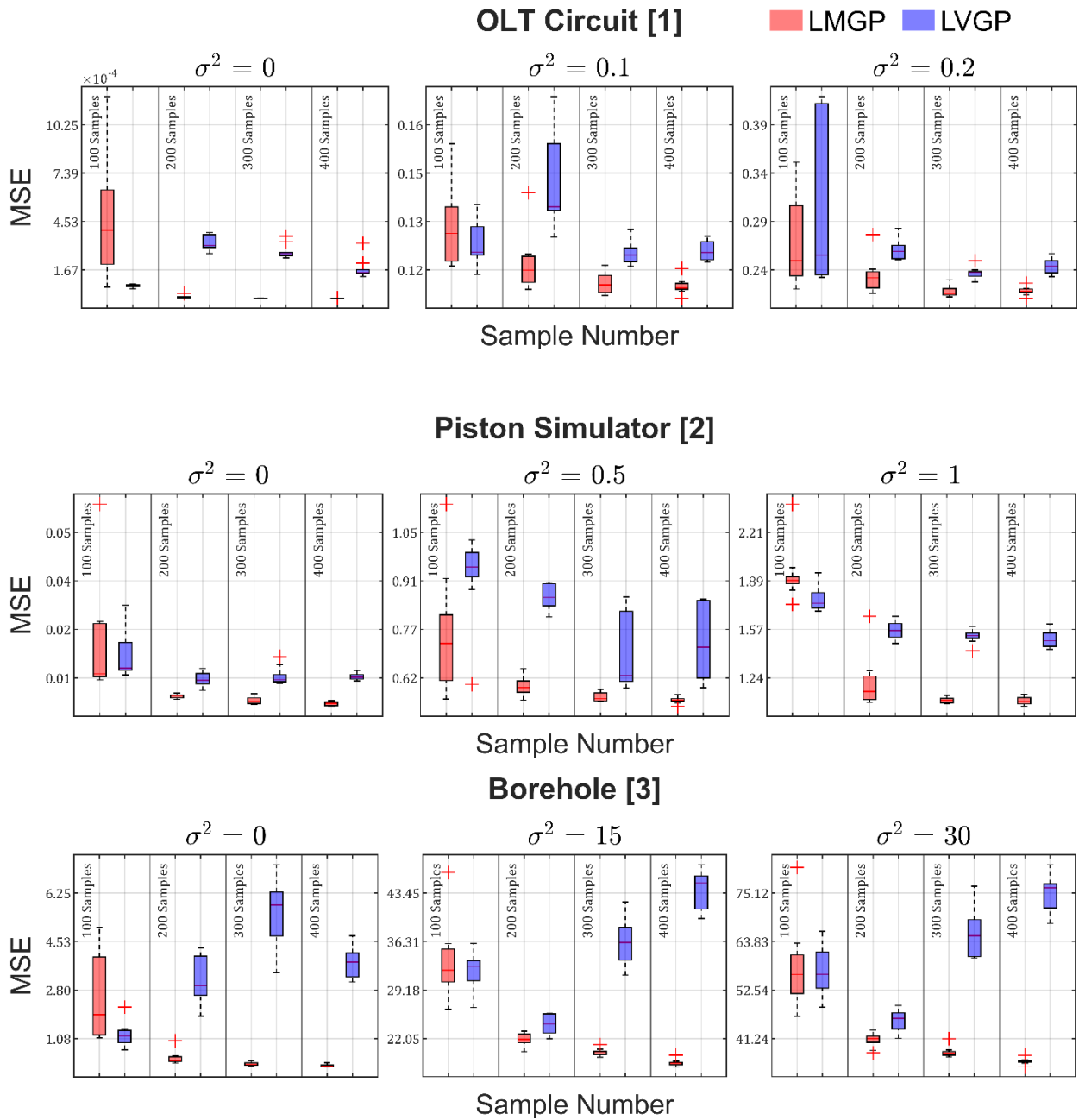

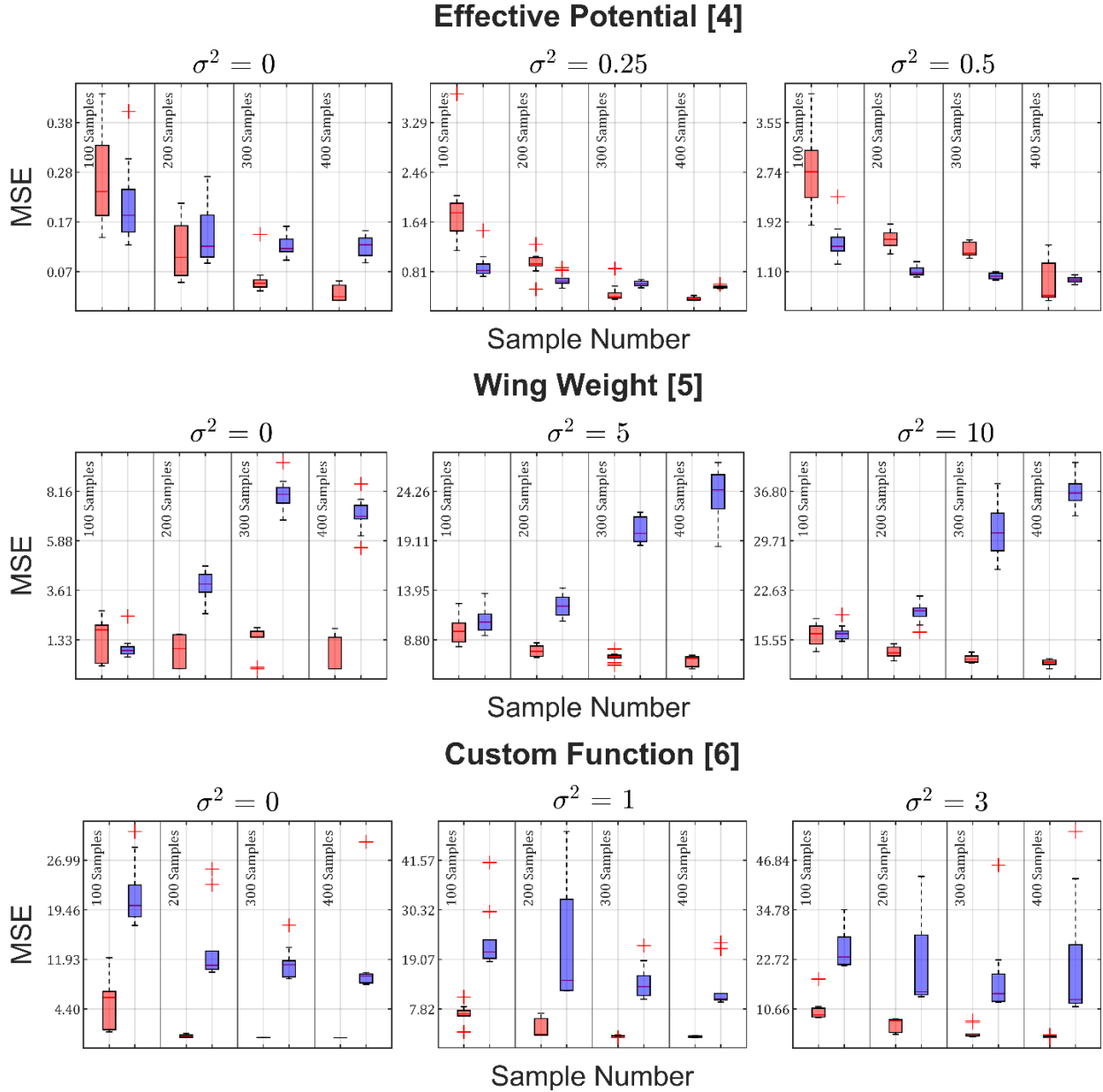

**Figure 4 Results on analytical functions:** We compare LMGP to LVGP across six different analytical functions. For each case, LMGP and LVGP are fitted to datasets of sizes $100, 200, 300$, and $400$ with three different noise levels (one noise level being $0$ and the other two depending on the range of the analytical function). $10,000$ noisy test data points are used to obtain MSE. The training and validation process are repeated $10$ times to account for randomness.

**Figure 5** shows the estimated noise variance via LVGP and LMGP across all the functions when $400$ training samples are used. From the boxplots, we see that LMGP more accurately estimates the noise variances in all cases except for the OLT circuit model (although the difference is very small in this case). When LVGP is less accurate, noise estimates are off quite noticeably.

We believe that the combination of embedding prior information on the categorical variable levels and using a single latent space for all categorical variables allows LMGP to discover a more general solution, making noise estimation via MLE easier.

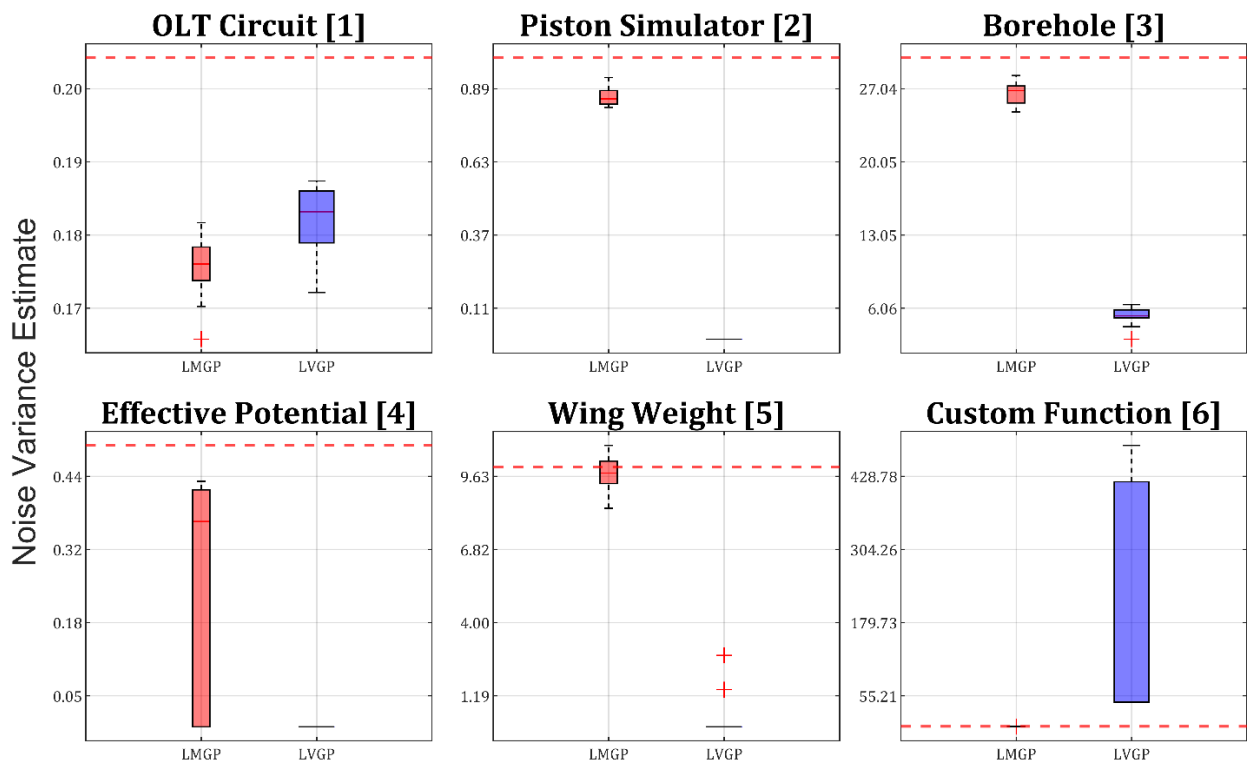

**Figure 5 Estimated noise variance:** Each time LVGP and LMGP are fitted, the noise variance is estimated using the training data. For each analytical function, 400 training samples are used with the injected noise levels indicated by the dashed red line in each subplot. It is clear that on average LMGP estimates the noise variance more accurately.

## 5.2  Real-World Datasets

In this section, we compare the performance of LMGP, LVGP, and standard GP across two datasets: the Boston housing dataset [61] and auto-MPG dataset [62]. In the former dataset, the goal is to predict the median housing prices in the suburbs of Boston in 1978. The dataset has 506 samples, 13 inputs, and one output. Since the output is capped at 50, the output of some samples is not trustworthy. Hence, we remove samples whose output is 50. As a result, the number of samples is reduced to 503. For LMGP and LVGP, CHAS (Charles River dummy variable) and RAD (index of accessibility to radial highways) are treated as categorical inputs. For GP, all inputs, all inputs are treated as numerical. The dataset is randomly split into 70% training and 30% validation. To account for randomness, the comparison test is performed 10 times.

In the auto-MPG dataset, the goal is to predict the MPG of various cars based on 8 inputs. After removing samples with missing values, 392 samples are available. For LMGP and LVGP, the number of cylinders and origin (i.e., the country the car was built in) are treated as categorical inputs while all inputs are treated as numerical for GP. The dataset is randomly divided into 50% training and 50% validation, and the comparison test is performed 10 times.

**Table 6** summarizes the results which show that LMGP slightly outperforms LVGP. Both models perform better GP, especially with the auto-MPG dataset. LMGP and LVGP perform similarly in these two datasets for the following reasons: ($i$) Both datasets only have two categorical variables where none of them has many levels, ($ii$) the categorical variables have little interaction, and ($iii$) at least one of the categorical variables has little effect on the response. GP's poor performance on the auto-MPG dataset is likely because the input, origin, should not be treated as a numerical input while the other categorical variables have some justification for being treated as numerical features. We note that the performance of LMGP and LVGP is either better or comparable to that of state-of-the-art NNs fitted to these datasets [63-66]. Unlike NNs, however, neither LMGP nor LVGP require iterative adjustment of, e.g., architecture, learning rate, or epoch number. In other words, training LMGP and LVGP is much simpler on such a small to medium size dataset.

**Table 6 Results on real world datasets:** The performance of LMGP, LVGP, and GP are analyzed across two real-world datasets. For each dataset, 10 different permutations of training and validation subsets are used, and the mean ($\mu_{MSE}$) and standard deviation ($\sigma_{MSE}$) of mean squared error (MSE) is reported.

|  | **Boston Housing** | | **Auto-MPG** | |
| --- | --- | --- | --- | --- |
|  | $\mu_{MSE}$ | $\sigma_{MSE}$ | $\mu_{MSE}$ | $\sigma_{MSE}$ |
| **LMGP** | 7.166 | 1.060 | 7.934 | 1.520 |
| **LVGP** | 7.371 | 1.408 | 8.241 | 1.427 |
| **GP** | 8.900 | 1.847 | 19.416 | 22.845 |

## 5.3 LMGP for Variable-Length Categorical Inputs

In this section, we analyze the performance of LMGP for handling variable-length categorical inputs. To this end, we reuse the borehole and OLT circuit functions defined in **Table 4** with some modifications. In particular, we remove certain categorical variables based on the level of other categorical variables. As shown in **Table 7**, we remove $(i)$ the third categorical variable when the first two categorical variables are at their respective level one, $(ii)$ the second categorical variable when the first and third categorical variables are at their respective level two, and $(iii)$ the first categorical variable when the last two categorical variables are at their respective level three. Regarding the underlying function, when the categorical variable is $NaN$, we set the variable to a numerical value as if it were another level (see **Table 8**). Note that this value is unknown to LMGP and not used in any way during the training.

**Table 7 Combinations of levels:** Listed are all cases, for the borehole and OLT circuit model, when one of the categorical variables is not an input (i.e., is NaN) given the level of the other categorical variables.

| Levels of Cat. Variable 1 $T_l$ (borehole) and $R_{b1}$ (OLT) | Levels of Cat. Variable 2 $L$ (borehole) and $R_f$ (OLT) | Levels of Cat. Variable 3 $K_w$ (borehole) and $\beta$ (OLT) |
|---|---|---|
| 1 | 1 | $NaN$ |
| 2 | $NaN$ | 2 |
| $NaN$ | 3 | 3 |

**Table 8 Underlying numerical values:** When a categorical variable's level is set to $NaN$ (i.e., the cases listed in **Table 7**), the underlying analytical function simply sets the categorical variable to an underlying numerical value while LMGP treats the variable as if it is no longer an input.

| | **Borehole Model** | | | **OLT Circuit Model** | | |
|---|---|---|---|---|---|---|
| | $T_l$ | $L$ | $K_w$ | $R_{b1}$ | $R_f$ | $\beta$ |
| **Underlying Value** | 350 | 1100 | 8000 | 35 | 1 | 2 |

For each function, we randomly generate 400 training samples and 10,000 test samples following the data generation strategy described in Section 5.1. We add IID normal noise with different variances (see **Figure 6**) to both training and test data. We then fit LMGP to the training data and evaluate it on the test data. To account for randomness, the procedures are repeated 10 times. **Figure 6** summarizes the results and indicates that both strategies described in Section 4.2.3 have similar performance. In particular, they both achieve MSE on test data relatively close to the applied noise, implying that the "Zero" and "Random" approaches are not causing significant losses in prediction performance.

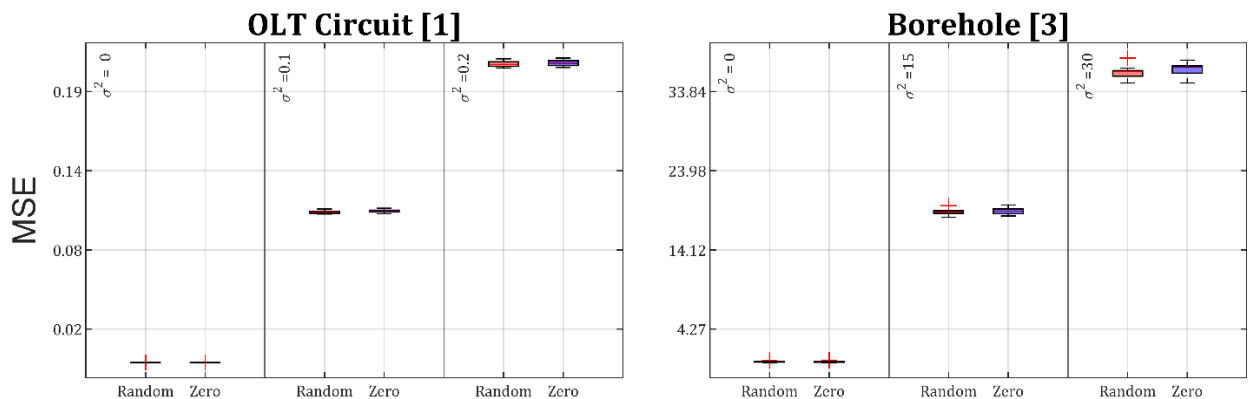

**Figure 6 Comparing LMGP methods for handling variable-length inputs:** Using the borehole and OLT circuit model, we compare predictive performance between two methods of handling variable-length inputs: The random and zero approach. The strategy of setting $NaN$ to a random value, $\chi$, and $0$ will be denoted as the zero and random approach, respectively. Both methods perform similarly with an MSE close to the injected noise variances.

### 5.4 Material Design with Bayesian Optimization

In this section, we apply LMGP to a material design problem previously studied in [67] where the goal is to use as few data points as possible (from a dataset of size $240$ [68]) to find the elements in the family of $M_2AX$ compounds that maximize bulk modulus. These compounds are nanolaminate ternary alloys that exhibit many of the beneficial properties of both ceramic and metallic materials and hence are appealing for many technological applications [68-70]. The family of $M_2AX$ compounds have three building blocks that can take on different elements: an early transition metal $M = \{Sc, Ti, V, Cr, Zr, Nb, Mo, Hf, Ta, W\}$, a main group element $A = \{Al, Si, P, S, Ga, Ge, As, Cd, In, Sn, Tl, Pb\}$, and either carbon or nitrogen $X = \{C, N\}$.

A direct strategy for finding the optimal compound is to compute the bulk modulus for all the $10 \times 12 \times 2 = 240$ candidates via density functional theory (DFT). However, this approach is suboptimal because DFT is computationally expensive. An alternative strategy is to use Bayesian optimization (BO) to discover the optimum compound by only obtaining the modulus of some of the candidates via DFT. A generic BO framework starts by fitting a probabilistic predictive model to an initial training data. Then, this model is used in an acquisition function that balances exploration and exploitation to identify the next candidate that must be evaluated and added to the training data. This three-step iterative process (that consists of model training, evaluation of acquisition function, and updating the training data) continues until the convergence criterion is met (e.g., resources are exhausted).

In BO, the user chooses the acquisition function and model type [71-74]. In this paper, we use expected improvement (EI) for the acquisition function defined as:

$$EI(\boldsymbol{x}) = E[max(y_{max} - \hat{y}(\boldsymbol{x}), 0)], \tag{13}$$

where $E[\cdot]$ denotes the expectation operator, $y_{max}$ is the best candidate in the current training dataset, and $\hat{y}(\boldsymbol{x})$ is the prediction for candidate $\boldsymbol{x}$. If the model's prediction has a normal distribution with mean $\mu(\boldsymbol{x})$ and variance $\sigma(\boldsymbol{x})$, Eq. (13) takes on the following closed form formula:

$$EI(\boldsymbol{x}) = (y_{max} - \mu(\boldsymbol{x}))\Phi\left(\frac{y_{max} - \mu(\boldsymbol{x})}{\sigma(\boldsymbol{x})}\right) + \sigma(\boldsymbol{x})\phi\left(\frac{y_{max} - \mu(\boldsymbol{x})}{\sigma(\boldsymbol{x})}\right), \tag{14}$$

where $\Phi(\cdot)$ and $\phi(\cdot)$ denote, respectively, the cumulative distribution and probability density functions of the standard normal distribution, $\mathcal{N}(\mu = 0, \sigma = 1)$.

Choosing the type of the predictive model is a challenge for this design optimization problem because all inputs are categorical where each categorical variable corresponds to a site ($M, A$, or $X$) and the levels represent potential elements for each respective site (e.g., $C$ or $N$ for site $X$). The strategy adopted in [67] is to use domain knowledge to convert categorical variables into quantitative inputs which can then be used in a standard GP. In particular, in this strategy each element (e.g., $Sc, Al$, or $C$) is characterized with its orbital radii ($s$-, $p$-, and $d$-orbital radii for elements at site $M$ while $s$- and $p$-orbital radii for $A$ and $X$ sites[3]) which, in turn, converts a candidate compound into a $7D$ quantitative variable.

Unlike the strategy of [67], LMGP can be directly applied to the original dataset which eliminates the time-consuming and problem dependent feature engineering step. This is advantageous because the replaced numerical variables may not sufficiently represent the effects of changing an element in one of the three sites. To demonstrate this benefit, we compare standard GP (with the categorical variables replaced with the $7D$ numerical variables) to LMGP when they are used as the predictive model in BO. In particular, we exclude the compound with the largest bulk modulus from the original dataset and then start the BO with randomly selected 40 compounds. We continue taking samples from the original dataset, one by one as guided by the acquisition function in Eq. (14), until the best compound is found. To compare GP vs. LMGP, we record the number of additional samples that BO evaluates until convergence. To account for the randomness, we repeat this process 30 times where each time a unique set of initial compounds is used.

---

[3] It is unclear why $d$-orbital radius is not used for $A$ and $X$ sites.

**Figure 7** is a histogram of the number of additional samples needed before finding the optimal compound which indicates that, on average, LMGP and GP require sampling 16.38 and 18.13 additional compounds, respectively. Thus, LMGP is more likely to find the optimal compound earlier than standard GP. We believe this is because the numerical features chosen for standard GP do not sufficiently capture the effects of switching elements for each site. LMGP does not assume the underlying numerical variables are solely defined by the orbital radii and thus, it is more flexible. Furthermore, we emphasize that LMGP did not require domain knowledge to identify the underlying numerical variables. This eliminates the need for feature engineering and makes our strategy very desirable for materials design and analysis where the underlying numerical features are not even known by domain experts.

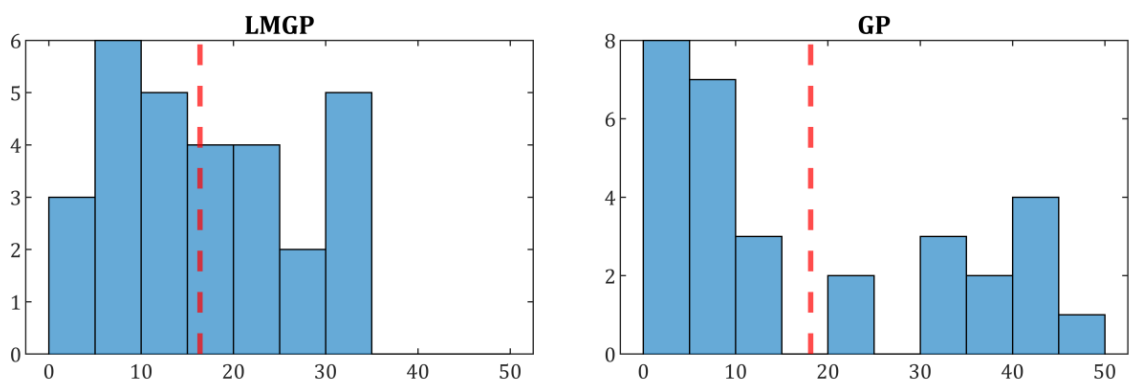

**Figure 7 Results of Bayesian optimization:** We compare the performance of LMGP and standard GP across 30 tests, each with a unique initial dataset of 40 compounds. The histograms indicate the number of additional compounds required to sample before finding the compound with the highest bulk modulus (smaller is better). The average number in each case is shown with the dashed vertical line.
2828

# 6 CONCLUSION

In this paper we introduced LMGPs which are extensions of GPs that can build surrogates with quantitative and qualitative inputs. As we showed, the main idea behind LMGPs is to learn a linear map that converts each combination of qualitative inputs to a point in a low-dimensional latent space. Since these latent points are endowed with an automatically learned distance measure, they can be directly used in any standard correlation function such as the Gaussian or Matérn.

We estimated the optimal linear map simultaneously with other hyperparameters by maximizing the Gaussian likelihood function. Alternatively, a Bayesian approach can be used to find the posterior distribution of LMGP's linear map. We have not pursued this in our studies yet.

By interpreting the linear map as an operator that projects all prior latent representations to their corresponding posteriors, we studied the effect of priors on LMGP. We showed that an informative prior consisting of grouped one-hot encoded inputs helps LMGP in building well-structured latent spaces and maximizes the performance on test data. Other types of priors may be more useful in applications where the fitted LMGP has to satisfy some physical constraints or where there is some prior knowledge on how qualitative inputs are related.

LMGPs can be interpreted as neural networks with certain architecture and activation functions. This interpretation opens up interesting possibilities that we will study in our future works. For instance, ($i$) the current linear map can be converted to a highly nonlinear one by adding hidden layers with nonlinear activation functions (e.g., sigmoid or swish activations), ($ii$) physics-informed LMGP can be built by infusing governing dynamics to the loss function, or ($iii$) very high-dimensional inputs-outputs can be handled by using convolutional layers.

# 7 ACKNOWLEDGEMENTS

We thank the original authors of LVGP for providing their Matlab code to compare the performance of LMGP to LVGP. We also thank Professor Daniel Apley for his detailed feedback on LMGP. This work was supported by the Advanced Research Projects Agency-Energy (ARPA-E), U.S. Department of Energy, under award number DE-AR0001209.

# 8 APPENDIX

## 8.1 Selection of LVGP and LMGP Parameters

For LVGP, the hyperparameter ranges are limited to: $\omega_i \in [-8, 3]$, $z_j^i(t_i) \in [-5, 5]$, and $\delta \in [0.1, 1E-10]$. For LMGP, the hyperparameter ranges are limited to: $\omega_i \in [-8, 3]$, $A_{i,j} \in [-1, 1]$, and $\delta \in [0.1, 1E-10]$. Both LMGP and LVGP estimate all the hyperparameters via a gradient-based optimization approach that starts the search via 12 initial, randomly selected points.

## 8.2 Underlying Numerical Values

In this section, we list the underlying numerical value associated with the different levels of each categorical variable for the analytical functions introduced in Section 5.1.

**Table 9 Underlying Numerical Values:** OLT Circuit Model [1]

| Categorical Variable Level | $R_{b1}$ | $R_f$ | $R_f$ |
|---|---|---|---|
| 1 | 25 | 0.5 | 1 |
| 2 | 32.5 | 2 | 4 |
| 3 | 40 | 3 | 5 |

**Table 10 Underlying Numerical Values:** Piston Simulator Model [2]

| Categorical Variable Level | $M$ | $S$ | $V_0$ |
|---|---|---|---|
| 1 | 30 | 0.005 | 0.002 |
| 2 | 40 | 1 | 0.4 |
| 3 | 50 | 2 | 1 |

**Table 11 Underlying Numerical Values:** Borehole Model [3]

| Categorical Variable Level | $T_l$ | $L$ | $K_w$ |
|---|---|---|---|
| 1 | 10 | 1000 | 6000 |
| 2 | 30 | 1400 | 10000 |
| 3 | 100 | 2000 | 12000 |
| 4 | 200 | $N/A$ | $N/A$ |
| 5 | 500 | $N/A$ | $N/A$ |

**Table 12 Underlying Numerical Values:** Effective Potential Model [4]

| Categorical Variable Level | $x_7$ | $x_8$ | $x_9$ | $x_{10}$ |
|---|---|---|---|---|
| 1 | 0.1 | 1 | 5 | 0.01 |
| 2 | 0.25 | 2 | 10 | 0.02 |
| 3 | 0.7 | 4 | 12.5 | 0.1 |
| 4 | 0.8 | 9 | 25 | 0.3 |
| 5 | 1 | 10 | 30 | 0.5 |

**Table 13 Underlying Numerical Values:** Wing Weight Model [5]

| Categorical Variable Level | $S_w$ | $W_{fw}$ | $t_c$ | $W_{dg}$ |
|---|---|---|---|---|
| 1 | 150 | 220 | 0.08 | 1700 |
| 2 | 180 | 250 | 0.12 | 2000 |
| 3 | 200 | 300 | 0.18 | 2500 |

**Table 14 Underlying Numerical Values:** Custom Function [6]

| Categorical Variable Level | $x_3$ | $x_4$ | $x_8$ |
|---|---|---|---|
| 1 | 0 | 0 | 0 |
| 2 | 0.1 | 0.2 | 0.4 |
| 3 | 0.3 | 0.7 | 1 |
| 4 | 0.6 | 1 | $N/A$ |
| 5 | 0.7 | $N/A$ | $N/A$ |
| 6 | 1 | $N/A$ | $N/A$ |